\documentclass[runningheads]{llncs}
\usepackage[T1]{fontenc}
\usepackage{graphicx}
\usepackage{algorithm}
\usepackage{algpseudocode}
\usepackage{amssymb}
\usepackage{graphicx}
\usepackage[utf8]{inputenc}
\usepackage{import}
\usepackage{listings}
\usepackage[numbers, sectionbib]{natbib}
\usepackage{hyperref} 
\usepackage{url}
\usepackage[dvipsnames]{xcolor}
\usepackage{multirow}
\usepackage{enumitem}
\usepackage{booktabs}
\usepackage{orcidlink}

\usepackage{fancyhdr}
\renewcommand{\subsubsection}[1]{\paragraph{\normalfont\normalsize\bfseries\textbf{#1}}}

\definecolor{editorOcher}{rgb}{1, 0.5, 0}
\definecolor{editorGreen}{rgb}{0, 0.5, 0}

\lstdefinelanguage{JavaScript}{
  keywords={const, let, typeof, new, true, false, catch, function, return, null, catch, switch, var, val, if, in, while, do, else, case, break},
  keywordstyle=\color{blue}\bfseries,
  ndkeywords={class, export, boolean, throw, implements, import, this},
  ndkeywordstyle=\color{red}\bfseries,
  identifierstyle=\color{black},
  stringstyle=\color{editorOcher}\ttfamily,
  commentstyle=\color{editorGreen}\ttfamily,
  sensitive=false,
  comment=[l]{//},
  morecomment=[s]{/*}{*/},
  morestring=[b]',
  morestring=[b]"
}

\begin{NoHyper}
\def\thefootnote{*}\footnotetext{The first two authors contributed equally to this work.}%
\def\thefootnote{\dag}\footnotetext{Email for correspondence.}%
\addtocounter{footnote}{-1}%
\end{NoHyper}

\begin{document}
\title{Learning Type Inference for\\ Enhanced Dataflow Analysis}
\titlerunning{Type Inference for Enhanced Dataflow Analysis}
%
\author{Lukas Seidel\thefootnote{*}\inst{1}$^\&$\inst{3}\orcidlink{0009-0006-8190-0300} \and
Sedick David {Baker Effendi}\thefootnote{*}\inst{2}$^\&$\inst{4}\orcidlink{0000-0002-4942-626X} \and
Xavier Pinho\inst{1} \and
Konrad Rieck\inst{3} \and
Brink van der Merwe\inst{2} \and
Fabian Yamaguchi\inst{1}$^\&$\inst{2}}
%

\institute{QwietAI, San Jose, USA \\
\email{jlseidel@qwiet.ai} \and
Stellenbosch University, Stellenbosch, South Africa \\
\email{dbe@sun.ac.za}\thefootnote{\dag} \and
Technische Universität Berlin, Berlin, Germany \and
Whirly Labs, Cape Town, South Africa}
\maketitle              
\begin{abstract}

    Statically analyzing dynamically-typed code is a challenging endeavor, as even seemingly trivial tasks such as determining the targets of procedure calls are non-trivial without knowing the types of objects at compile time. 
    Addressing this challenge, \emph{gradual typing} is increasingly added to dynamically-typed languages, a prominent example being TypeScript that introduces static typing to JavaScript.
    Gradual typing improves the developer's ability to verify program behavior, contributing to robust, secure and debuggable programs.
    In practice, however, users only sparsely annotate types directly.
    At the same time, conventional type inference faces performance-related challenges as program size grows.
    Statistical techniques based on machine learning offer faster inference, but although recent approaches demonstrate overall improved accuracy, they still perform significantly worse on user-defined types than on the most common built-in types.
    Limiting their real-world usefulness even more, they rarely integrate with user-facing applications. 
    
    We propose CodeTIDAL5, a Transformer-based model trained to reliably predict type annotations.
    For effective result retrieval and re-integration, we extract usage slices from a program's code property graph.
    Comparing our approach against recent neural type inference systems, our model outperforms the current state-of-the-art by 7.85\% on the ManyTypes4TypeScript benchmark, achieving 71.27\% accuracy overall.
    Furthermore, we present JoernTI, an integration of our approach into Joern, an open source static analysis tool, and demonstrate that the analysis benefits from the additional type information. 
    As our model allows for fast inference times even on commodity CPUs, making our system available through Joern leads to high accessibility and facilitates security research.

    \keywords{Type Inference, Representation Learning, Static Analysis, Static Taint Tracking, Dataflow Analysis}

\end{abstract}

\begin{center}
\begin{small}
    \textcolor{darkgray}{This is a preprint version of an article to be published at the \textit{28th European Symposium on Research in Computer Security} (ESORICS '23).}
\end{small}
\end{center}
\newpage

\section{Introduction}
\label{sec:intro}

Dynamically typed languages are continually rising in popularity, with JavaScript and Python consistently in the top 5 languages learned by developers~\cite{octoverse}.
While easier to learn and modify, these loosely typed languages do not benefit from compile-time error detection, optimizations, and IDE support, as statically typed languages do~\cite{park2014javascript, hanenberg2014empirical, gao2017type}. 
Statically typed languages often suffer from much of the same shortcomings when library dependencies are not available and whole-program analysis is not an option~\cite{dagenais2008enabling}, with hidden type declarations and class members leading to incomplete intermediate representations. 
If accessing all the dependencies of a project is not an option, this issue is then compounded by the growing catalog of open-source libraries and developers' inclination to make use of third-party code~\cite{xu2020reinventing}.
Overall, this incomplete view of the typing system of a given code base presents static program analysis with substantial challenges.
Traditional Static Application Security Testing (SAST) tools rely on type information to identify the attacker surface and sensitive data, and may fall short when they fail to infer it for given variables.
When analyzing the flow of data throughout a program, e.g., in order to find paths leading to XSS or the leak of sensitive information, the tool needs to be able to verify which method is implemented by what class, in order to track that flow recursively.
If now the method belongs to an external class, but it is unknown what type from what library or if there are multiple local classes implementing the same interface in different ways, the SAST tool is faced with a problem if the exact type of a variable is missing.
Statistical type inference can assist in deriving type suggestions for objects in dynamically typed languages, and a variety of machine learning and rule-based approaches were explored in recent years~\cite{hell18, lambdanet, typebert}.
But although these novel approaches were able to offer improvements time after time, implementation of such tools in real-world applications, actually used by developers, is heavily lacking. 
With only very recent work looking at integrations into development environments~\cite{voruganti2022flextype} and not a single published use case for program security analysis, this drastically increases the barrier to entry.

In this paper, we propose a new, holistic type inference system for dynamic languages:
We combine automated extraction of relevant data using an open source multi-language code analysis platform, Joern~\cite{joern}, with a Transformer model for type inference following an encoder-decoder architecture. 
Our approach is implemented for JavaScript, which can use the inferred type information in subsequent static analysis tasks.
Our proposed type inference mechanism outperforms the current state-of-the-art in machine-learning-based type inference by up to 7.85\%.
Even where it is not significantly better overall, our system still shows favorable performance where inference is most needed, i.e., on custom user-defined types.
Moreover, by training the model on annotated TypeScript code and then running inference as part of Joern's analysis on JavaScript, we show that re-integrating inferred types into the static analysis engine can improve dataflow recovery, allowing a developer to analyze their program in a more complete manner. 
  
In conclusion, our work makes the following contributions, advancing the current state of probabilistic type inference of dynamic languages for static analysis:
\begin{enumerate}
    \item We propose \emph{CodeTIDAL5}: A {CodeT5+} model for \textbf{T}ype \textbf{I}nference for enhanced \textbf{D}ataflow \textbf{A}nalysis via \textbf{L}anguage modeling. 
    Our machine learning model for type inference in JavaScript achieves above-state-of-the-art accuracy with a generative approach and increased context sizes. 
    Its open type vocabulary allows the model to provide useful hints even on types it has not seen during training.
    
    \item We present \emph{JoernTI}, an integrated type inference subsystem for Joern, and demonstrate the benefits in practice. 
    The module queries CodeTIDAL5 to then re-integrate inferred types into the internal code representation for subsequent analysis steps. 
    We show that additional type annotation coverage may lead to better dataflow analysis results. 

    \item We implement usage slicing in Joern: based on abstract syntax information, 
    our system generates usage vectors that can be used to efficiently locate and re-integrate inferred type information.
    We publish a dataset containing usage slices of 300 000 open source projects: \\\url{https://doi.org/10.5281/zenodo.8321614}. 
\end{enumerate}

\section{Background}

\label{sec:backgrnd}

\subsection{Code Property Graphs and Code Analysis}
\label{subsec:cpg}

The Code Property Graph (CPG) is a versatile data structure that combines multiple traditional program representations to form a holistic representation of a program's source code. 
Yamaguchi et al.~\cite{cpg} combine Abstract Syntax Trees (ASTs), Control Flow Graphs (CFGs), and Program Dependence Graphs (PDGs) into a single structure in order to capture relationships between syntax, control flow, and data dependencies in a unified view. 
CPGs may also incorporate symbol and type information, enabling comprehensive interprocedural and whole-program analysis. 
Moreover, many dataflow tasks can be solved with graph traversal, e.g., IFDS (interprocedural, finite, distributive, subset)~\cite{reps, reps95}, hence, CPGs offer a natural foundation for complex program analysis. 

Joern~\cite{joern} is the artifact of Yamaguchi et al.'s CPG associated research and
supports the analysis of various programming languages. 
Joern provides a Scala-based, domain-specific querying language to query against CPGs that allows security analysts to write succinct, high-level queries to traverse the CPG and match graph patterns indicative of potential vulnerabilities or violations of coding best practices. 
A major advantage is that Joern performs fuzzy parsing, making it robust to missing code and useful for partial program analysis tasks, not requiring a complete compilation environment.
By providing a unified representation of the code structure and its semantics, CPGs also enable effective tracing of data- and control-flow through an application, helping to identify potentially problematic semantic patterns that more limited representations might miss. 
In order to trace the flow of data in an inter-procedural way, techniques such as taint tracking are used to model the impact of input parameters on the returned values of a function~\cite{dfanalysis}.
Recursively, knowledge of dataflow for all used functions is required.
To this end, at least a partially-qualified method name is necessary to tag the correct function as a sensitive source or sink during taint analysis or in re-using the correct method summaries.
Ultimately, knowledge of an object's runtime type will reduce the number of procedures to consider and improve both the precision of subsequent analysis.

\subsection{Statistical Type Inference}
In contrast to strongly typed languages such as Java or Rust, \textit{gradually typed} languages do not strictly require the programmer to provide annotations for all variables but statically check types at compile-time if provided.
While type annotations and static type checking are optional in these gradually typed languages, such as TypeScript, they have a significant impact on the code quality~\cite{park2014javascript, tont}.
Therefore, inferring types is an important task in gradually or even dynamically typed languages, where types are not explicitly declared at all and might even change over the lifetime of a variable.
Automated type inference can reduce the manual effort required to add annotations to existing TypeScript code or to provide type hints to otherwise completely untyped JavaScript in a static analysis context.
This is possible, as JavaScript syntax is completely compatible with TypeScript and even implementations of popular open source libraries are re-used between the two languages, leading to the usage of the same classes or interfaces.
The TypeScript compiler, for example, can transpile JavaScript to TypeScript, enabling static type checking and its benefits for JavaScript code bases~\cite{understandingts}.
It also introduces rudimentary rule-based type inference, which unfortunately does not add a lot of value for complex or user-defined types, leaving many variables labelled as the ambiguous \texttt{any} type~\cite{lambdanet}.

The domain of statistical or \emph{neural type inference} has become well established as of late in an effort to statically recover type information for variables in application code using machine learning~\cite{hell18, lambdanet, pan20, pradel2020typewriter, ye20, typebert}. 
These models range from those that use simple token sequences with text models without constraints, such as DeepTyper~\cite{hell18} and TypeBert~\cite{typebert}, or with constraints, such as OptTyper~\cite{pan20} and TypeWriter~\cite{pradel2020typewriter}, to Graph Neural Network (GNN) based models that account for syntactic and semantic relations between code entities such as LambdaNet~\cite{lambdanet} and the R-GNN family \cite{ye20}.
While some work tries to introduce additional information by noting the constraints of the type in the context of the target object's usage~\cite{pan20, pradel2020typewriter}, we find that insufficient data is being learned on precise contextual hints such as how exactly the target object interacts within the surrounding procedure. 
Jesse et al.~\cite{typebert} address the type inference problem by using a much larger, BERT-style~\cite{bert} model and more training data than previous work, but this runs the risk of impractically large inference times and resource demand.

Although mentioned work progressed the academic state-of-the-art in type inference rapidly in recent years, virtually no work directly explores means to integrate their approaches into enhancing downstream tasks such as taint analysis.
This heavily restricts usability and limits adaptation, since even if proposed solutions are significantly better than readily available systems such as the TypeScript compiler's inference capabilities, developers do not have access to them outside a very limited set of tools such as FlexTyper~\cite{voruganti2022flextype}.

\subsection{Large Language Models}
In recent years, Large Language Models (LLMs), machine learning models with millions up to hundreds of billions trainable parameters, have introduced substantial improvements in many Natural Language Problem (NLP) tasks~\cite{peters-etal-2018-deep, generativepre, gpt3, bert}.
Training is performed on natural language text input in a self-supervised manner, i.e., learning objectives, so-called labels, for a given input are automatically retrieved from the raw input.
No manual labelling effort is required for generative pre-training performed in such a way, facilitating training on vast amounts of data.

Another crucial building block in the success of LLMs are \emph{Transformers}~\cite{transformers}.
At a Transformer's core, the self-attention mechanism allows the model to weigh the relevance of each element in the input sequence when processing a particular element, capturing the dependencies between elements regardless of their distance from each other.
Unlike recurrent neural networks, such as the widely adopted LSTMs~\cite{lstm} that process data sequentially, Transformers can process all data points in the input sequence simultaneously. 
This allows for high parallelization and makes Transformers particularly well-suited for modern hardware accelerators such as GPUs, leading to significant speed-ups in training time and facilitating scalability.
Two flavors of Transformer models are predominant:
\textit{GPT}~\cite{generativepre} is trained to generate the most probable next word, given a sequence of preceding words.
\textit{BERT}~\cite{bert} is a bidirectional approach, not only considering previous context but also subsequent words.
Where GPT is a decoder-only architecture, focusing on generating new text from its learned embeddings, BERT-style models typically only consist of encoders, whose main task is to capture contextual information as a continuous representation. 
GPT is commonly used for language modeling, e.g., for programming code synthesis~\cite{codegen}.  
BERT on the other hand excels at understanding context and is used for tasks such as sentiment analysis.

We base our approach on \textit{T5}~\cite{t5}, a unified encoder-decoder architecture introduced by Google. 
T5 was created to explore the boundaries of transfer learning by converting all problems into text-to-text, allowing it to be flexible to a number of NLP tasks. 
More specifically, we make use of Salesforce's CodeT5+~\cite{codet5p, wang2021codet5}, considering its identifier-aware pre-training showing comprehensive semantic understanding on the CodeXGLUE \cite{codexglue-paper} programming tasks.
A variety of pre-trained LLMs is readily available as open source, e.g., at Hugging Face~\cite{hfpt}.


\section{Motivation}
\label{sec:motivation}


Despite the rapid advancements in the precision of current state-of-the-art neural type inference models' ability to infer object types, there are still a number of limitations prohibiting their widespread adoption.
TypeBert's~\cite{typebert} main premise is to reduce the so-called inductive bias of inputs to a machine learning model.
Instead of computing sophisticated input formats, involving hand-crafted type constraint systems or explicitly encoding relations between syntax and semantic~\cite{lambdanet}, the authors scale up the number of parameters and the amount of raw data and let the model learn representations on their own.
Although this approach indeed works well, such token classification approaches still require careful processing of the output, although they already take the most straightforward format as input, i.e., raw code snippets.
In order to make use of the results in another environment, variable names, declarations and their positions in code files must be carefully matched against the respective token vectors.
No available system has the ability to be queried for the type of a single object in a clearly defined format, limiting integration potentials into other platforms.
Therefore, broadly accessible type inference systems that are actually available where developers need them, e.g., the inference passes of the TypeScript compiler, lag behind academic progress by years.

We identify the following main roadblocks as the primary reasons for lack of adoption of current state-of-the-art probabilistic type inference systems:
\begin{description}
    \item \textbf{[R1] Limited Performance on User-Defined Types.} Although increasing the overall type annotation coverage is definitely a laudable goal, annotations for built-in types are rarely the critical piece of information missing. In the place where type annotations are most crucial, where users are working with user-defined classes found either, are where available approaches are falling short. At the same time, precisely these types are necessary for improving the precision of downstream static analysis starts.
    \item \textbf{[R2] High-Effort Setups.} Many of the recently proposed solutions require complex setups to compile and run their application, and even if in a runnable state, the systems only handle highly specific in- and output formats requiring manual effort to receive applicable results.
    \item \textbf{[R3] No Integration with Existing Platforms.} Even the most usable of the available systems require a manual transfer from the type inference application to the user's IDE or a code analysis platform. This heavily restricts usability and adaptation, as the introduced overhead is unacceptable to most every-day users.
\end{description}
While subsets of these roadblocks were partially addressed by prior work,
to the best of our knowledge no approach tackled all of them systematically.
Consequently, with this work we aim to overcome them in a unified and highly usable system.
\section{Design}
\label{sec:design}
In the following, we discuss design decisions and core aspects of our usage slicing procedure, the machine learning model we train for type inference and its integration in Joern.

\subsection{Code Property Graph Usage Slicing}
\label{sec:slicedesign}

Where program slicing in general is the act of reducing a program to a subset of its information relevant for the current task, code property graph usage slicing reduces a full CPG of a given code base to a subset of useful nodes.  
More specifically, the goal of CPG \textit{usage slicing} is to extract meaningful information describing how an instantiated object interacts within a procedure while being robust to missing type and data-flow information. This may be seen as similar to TypeT5's \cite{typet5} \emph{usage graph}, but we omit ``potential usages'', remain intraprocedurally bound while still including variable usages captured by closures.
We compute CPG usage slices using abstract syntax information (as opposed to full-blown dataflow slicing \cite{hor90}), including dynamic calls invoked from this object or calls the current object is an argument for. Supplementary call graph and type information is used when available.
The method source code and these usage vectors build the foundation of our machine learning model.
It is important to note that a slice starts at some kind of definition of a variable and ends before any reassignment, so that we can guarantee a slice is representative of a single type.

Our approach rests on two concepts: variables and types. Variables are defined by a set $S$, and types by a set $T$.
Consider all types to make up the set $T$, and specific types form subsets of $T$, where variables at a specific point in a program may only belong to a single type $t \in T$.
Examples of primitive types are integer ($\mathbb{Z}$) and boolean ($\mathbb{B}$), as they represent a single word of data. There may exist a type \texttt{Foo} that defines integer and boolean members, which would imply these primitive types are a subset of \texttt{Foo}. Similarly, a collection of one or more characters ($\Sigma$) may fall in the string type set ($\Sigma*$).
We use $\varnothing$ to denote the \texttt{null} type, but also include types such as \texttt{void} in Java, \texttt{undefined} in JavaScript, \texttt{None} in Python, etc. which may appear in method signatures or type arguments. To denote the \emph{set of all types}, we use $\mathbb{U}$ (similar to \texttt{any} in JavaScript, \texttt{Object} in Java, etc.)

With these concepts, we can now introduce our definition set $D$. A definition $d \in D$ can be described as the tuple $(s,t)$ where $s \in S, t \in T$. This is more formally defined in Equation \ref{eq:def}, and is similar to what is used in \cite{pan20}.

\begin{equation}
    (s, t) =  d \in D
    \label{eq:def}
\end{equation}
We now present Listing \ref{lst:codeAppendix} as our running example and to illustrate the concepts described in this section. Within the scope of the closure at line $4$, we see three referenced variables from within the function: \texttt{req}, \texttt{res}, and \texttt{params}. From outside the scope of this closure, we capture \texttt{documentClient}. This slice excludes the implementation of the child closure at line $6$, but will trace usages of \texttt{documentClient} as it is captured in this procedure.


\begin{lstlisting}[
    language=JavaScript,
    caption={A JavaScript example of a generic HTTP request handler performing a query to a DynamoDB instance. A flow exists from \texttt{req}, over \texttt{params} into \texttt{query()}.},
    label={lst:codeAppendix}
]
const db = require("db.js");
const documentClient = db.documentClient;

const handler = (req, res) => {
  const params = req.body.params;
  documentClient.query(params, function(err, data) {
    if (err) console.log(err);
    else console.log(data);
  });
};

export default handler;
\end{lstlisting}

As is, without type information, the associated types of our variables would be the \texttt{any} type ($\Sigma*$). This is commonly how slices appear in practice. With type information, our variables become (\texttt{documentClient}, \texttt{DocumentClient}), (\texttt{req}, \texttt{NextApiRequest}), (\texttt{res}, \texttt{NextApiResponse}), (\texttt{params}, \texttt{Object}), (\texttt{err}, \texttt{Error}), and (\texttt{data}, \texttt{Object}). TypeScript programs enable us to populate these tuples with type information more often.

Finally, the usage slice is a 3-tuple consisting of two definitions and a set of calls. The definitions are (i) the object call, identifier, or literal that defines the data of the target object, named $d_{\textrm{\small def}}$, and (ii), the target object of the classification, named $d_{\textrm{\small tgt}}$. If $d_{\textrm{\small def}}$ is a parameter, then $d_{\textrm{\small tgt}}$. The calls are those which $d_{\textrm{\small tgt}}$ invokes or is a parameter to.

\subsection{Machine Learning Model}
\label{sec:transdesign}
\subsubsection{Architecture}
Conceptually, our model is meant to learn from semantic relationships, e.g., variable naming conventions and class names.
As we deliberately aim to deduce clues on an object's type from the semantic principles of how developers name variables or a class's methods, we opt for an encoder-decoder Transformer model for contextual text processing and generation.

The proposed machine learning model for type inference, hereinafter referred to as \textit{CodeTIDAL5}, is based on Saleforce's {CodeT5}+~\cite{codet5p} models.
We use the smallest version of the model family with 220M trainable parameters.
CodeT5+ is an encoder-decoder model trained on various uni- and bimodal pre-training objectives:
To initialize the model, it is first pre-trained with causal language modeling and denoising objectives on large amounts of source code from Github.
As a result, the model learns a robust semantic model for programming code and its relation to natural language.
Other pre-training objectives include text-to-code causal language modeling, a bimodal task in which both encoder and decoder are activated, generating code snippets from natural language descriptions or vice-versa.
The authors argue, that this type of objective is effective in closing the pretrain-finetune gap for downstream tasks such as code summarization~\cite{wang2021codet5}.
CodeT5+ achieves state-of-the-art results on programming tasks such as natural language code search, in which a model needs to find the most semantically
related code from a natural description, and code completion.
We model the type inference problem as a sequence-to-sequence task, tagging variable locations of interest in raw code snippets and letting the model generate token-based type predictions per tag.

We make use of Hugging Face's \texttt{transformers} library~\cite{hf-transformers} for model implementation and training.
The library implements various state-of-the-art optimizations and best practices, e.g., FlashAttention~\cite{flashattention} that introduces substantial speedups and memory reductions to Transformers.

\subsubsection{Input and Output Representations}
During inference, we extract usage information from our usage slices to accurately match objects with their occurrences in raw source code.
Given a JavaScript or TypeScript code base, comprehensive usage slices can be extracted with Joern's \texttt{joern-slice} capability.
We annotate an object's declaration and usage locations in the source code of a given function for which we want to retrieve type suggestions with special token tags, signaling relevance of a certain variable to the model.
This annotated code snippet is subsequently prefixed with a task description.
We finally transform the input into its numerical representation, using CodeT5's tokenizer.

The model generates its output token-by-token, responding with a list of tag-to-prediction assignments of the form \texttt{"<extra\_id\_0> Array"}.
Code snippets without any tags receive \texttt{"No types to infer."} as a label during training.
An input context length of 512 tokens was used during training and evaluation, and output length is restricted to 128 tokens.
As opposed to BERT-style models with a token classification head on top, such as TypeBert~\cite{typebert}, this theoretically allows the model to produce type suggestions outside the corpus of types it has seen during training.
We investigate the occurrence and usefulness of such hallucinations in Section~\ref{sec:pa}.

\subsubsection{Training} For training and testing purposes, we make use of the 
ManyTypes\-4TypeScript dataset~\cite{mt4ts}.
The test split comprises 662 055 TypeScript functions with a total of 8 696 679 type annotations, featuring 50 000 different types.
After pre-processing and tokenization for the sequence-to-sequence task, we have 1 758 378 samples in the training set. 
We fine-tune CodeTIDAL5 for a total of 200k steps.

\subsection{Integration in Joern}
\label{sec:joernti}

At the core of the CPG is a language agnostic AST schema from which subsequent analysis and semantics are built upon. 
The first component of building a CPG is the \emph{language frontend} that uses a parser and abstracts the source code to the AST. 
Once the AST is complete, subsequent analysis is performed and the graph is then annotated with additional nodes and edges. 
These subsequent analyses are called ``passes'' and an example is the CFG pass which accepts the AST and annotates it with CFG edges. 
Similarly, data-dependence depends on the CFG as input, so once the CFG pass is complete, the control-dependence and reaching-definitions passes are run to construct the intraprocedural CPG.

\begin{figure}[tp]
    \centering
    \includegraphics[width=\linewidth]{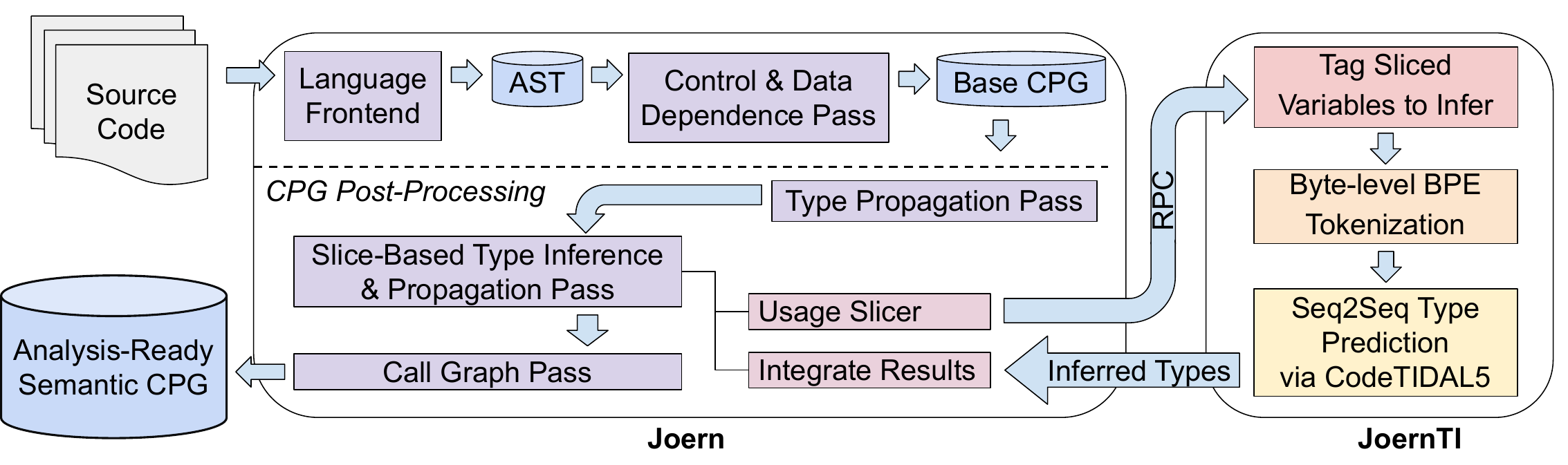}
    \caption{High-level overview of Joern submodules and integration of our type inference system.}
    \label{fig:joernti}
\end{figure}

We implement our analysis in a post-processing pass that accepts the intraprocedural CPG and extracts usage slices, which are then given to CodeTIDAL5 for type inference. The pipeline is illustrated by Figure \ref{fig:joernti}.

\subsubsection{Type Propagation Pass}
\label{sec:xtyperec}
The type propagation pass is a simple flow-insensitive type recovery algorithm whereby variables, parameters, and fields are associated with types they are assigned to or annotated with.
This information is kept in a map which is valid for the scope of the file. 
For example, if a variable \texttt{x = 1} is encountered, we associate \texttt{x} with an integer, but if we later see \texttt{x = "foo"} then we append the set of associated types with string. The result would be that \texttt{x = [integer, string]}. 
However, if at the end of the type propagation, we note that \texttt{x = [integer]} then we can be almost certain it is of that type. 
While this kind of algorithm is often run as a fixed-point calculation, this is run for a small fixed number of iterations to recover the majority of simple types quickly, while CodeTIDAL5 can recover the rest. 
This type propagation also gathers type hints for procedure parameters and return values from caller-callee pairs, as well as what occurs within the procedure body itself. 
This is also why we need to then propagate the inferred types once we receive our results from CodeTIDAL5, in order to make sure interprocedural usages reflect the inferred types. 

\subsubsection{JoernTI Server} We implement JoernTI, a Python package acting as a queryable server, offering access to slice-based type inference. 
We implement an enhanced type inference Joern pass which, after the type propagation pass, slices the program and sends a payload of usage slices with their corresponding scope's raw source code, called a ``program usage slice'', to the JoernTI server. 
The response produced by CodeTIDAL5 is a collection of variables and their inferred types. 
The inference results are then integrated into the CPG, where the new type information can be utilized downstream for building the call graph, source-sink tagging, and dataflow analysis. 
As the model may generate incorrect types, we implement a configurable look-up on the CPG to validate the suggestion based on simple type constraints. 
Using TypeScript's type declaration files (\texttt{d.ts}), one can load class definitions into Joern where inferred types can be checked against what the object invokes and which properties it accesses, e.g., \texttt{Request} has a \texttt{body} property but does not have a \texttt{connect()} method (unless the \texttt{.prototype} is modified). 
Examples of such files can readily be found in public repositories~\cite{deftyped}.
\section{Evaluation}
\label{sec:eval}
In the following, we compare the performance of CodeTIDAL5 against state-of-the-art probabilistic type inference systems for JavaScript/TypeScript. 
As a real-world evaluation, we use CodeTIDAL5 in partial-program analysis environments and validate the number of notable types recovered, especially those of which would be useful in taint analysis tasks.

\subsection{Type Inference Generalization}
\label{sec:benchmark}

We compare our approach against the following systems:
\begin{enumerate}[topsep=0pt]
    \item \textbf{LambdaNet~\cite{lambdanet}:} A robust baseline for machine-learning-based type inference systems, whose dataset was commonly re-used in subsequent work.
    \item \textbf{TypeBert~\cite{typebert}:} An early but very successful approach to address the domain, mainly as a Natural Language Modeling problem with Transformers. We use the author's Hugging Face implementation~\cite{typebert-hf}.
    \item \textbf{GCBert-4TS~\cite{mt4ts}:} A GraphCodeBERT~\cite{graphcodebert} base model fine-tuned on the ManyTypes4TypeScript dataset, currently leading the
    \\ 
    CodeXGLUE~\cite{codexglue-paper, codexglue-bench} benchmark for type inference. We again use the author's Hugging Face implementation~\cite{graphcodebert-hf}.
\end{enumerate}
\subsubsection{Datasets}
We evaluate all approaches on variables with user-provided types from the original LambdaNet (LN) dataset~\cite{lambdanet}, comprised of 60 TypeScript GitHub Repositories, and on the ManyTypes4Typescript (MT4TS) dataset~\cite{mt4ts}.
We use the MT4TS dataset as is.

For LambdaNet, we construct a high-quality dataset split.
We extract predictions from the openly accessible experiment data for the paper~\cite{ln-data}.
In accordance with our goal to use type inference for dataflow analysis, we remove uninteresting objects from the dataset.
First, we remove the non-expressive type annotations \texttt{Function} and \texttt{void} from the sample set.
Secondly, we only consider variables with usages besides assignments.
We also remove objects being part of class definitions, as we will infer types inside any function which uses or accesses such class members and, again, their occurrence purely as part of a definition is uninteresting from a dataflow point of view.
Furthermore, we aim for a more challenging dataset, better reflecting real-world use cases of type inference where already annotated types are of no interest:
While we use user-provided type annotations as a ground truth, we mask these annotations during inference for TypeBert, GCBert-4TS and CodeTIDAL5.
For this, we obfuscate all object instantiations where a type's name can be derived from the \texttt{new} call, as well as manual type annotations.

\subsubsection{Metrics}
We report Top-1 accuracy for the following type subsets, in accordance with previous work:
The \textbf{Top-100} subset includes the 100 most frequent built-in types. Examples are \texttt{String} or \texttt{Array}.
\textbf{User-Defined Types} are locally, i.e., within the same scope, defined classes or enums.
We derive this category from the original LN experiments, and hence do not report this metric for MT4TS.
As LambdaNet excludes external types from their prediction space, types from imported libraries are not considered, limiting the dataset's scope.

For evaluation on the LN dataset, for all approaches we perform greedy matching:
Specifically, we infer the type of a variable at multiple locations where it is used in the source, instead of only, e.g., at instantiation, and choose the highest confidence match in order to evaluate a label against the ground truth for a variable.
On MT4TS, we only consider exact matches per unique variable usage location.
In accordance with previous work, we exclude labels with the ambiguous \texttt{any} tag, and consider \texttt{UNK} predictions as incorrect.

\begin{table*}[tp]
    \centering 
    \caption[Evaluation Results for LambdaNet dataset]{Performance comparison of ML-based Type inference systems for TypeScript on the LambdaNet~\cite{lambdanet} dataset. Size in number of trainable parameters.}
    \label{tab:tseval}
    \begin{tabular}{l|ccc|c}
    \toprule
    Model & \multicolumn{3}{c| }{Top-1 Acc \%} & Size \\
    & \multicolumn{1}{c}{User-Defined} & \multicolumn{1}{c}{Top-100} & \multicolumn{1}{c|}{Overall} & \\
    \midrule
    LambdaNet & \multicolumn{1}{c}{46.89} & \multicolumn{1}{c}{64.79} & \multicolumn{1}{c |}{61.18} & N/A \\
    TypeBERT & \multicolumn{1}{c}{51.50} & \multicolumn{1}{c}{73.30} & \multicolumn{1}{c |}{68.90} & 360M \\
    GCBert-4TS & \multicolumn{1}{c}{46.89} & \multicolumn{1}{c}{\textbf{80.62}} & \multicolumn{1}{c |}{\textbf{73.81}} & 162M \\
    CodeTIDAL5 & \multicolumn{1}{c}{\textbf{53.20}} & \multicolumn{1}{c}{79.77} & \multicolumn{1}{c |}{73.61} & 220M \\
    \bottomrule
    \end{tabular}
\end{table*}

\subsubsection{Results}
Performance comparisons of CodeTIDAL5 against state-of-the-art neural inference systems for TypeScript are presented in Tables \ref{tab:tseval} and \ref{tab:tseval-mt4ts}.
On the more diverse and larger ManyTypes4TypeScript dataset, our approach achieves an overall improvement of 7.85\% in type prediction accuracy against the current best published model GCBert-4TS.
TypeBert's comparatively low accuracy is probably explained by the reduced context size. 
Where GCBert-4TS and CodeTIDAL5 see a context window of 512 code tokens at a time, TypeBert is trained to only process inputs of 256 tokens.
On all benchmarks, CodeTIDAL5 is at least on-par with GCBert-4TS, besting all other evaluated approaches by wide margins.
Although overall accuracy on the LN dataset did not show improvements, our approach's performance on user-defined types is significantly better.

\begin{table}[tp]
    \centering 
    \caption[Evaluation Results for Typescript]{Performance comparison on the ManyTypes4TypeScript dataset.}
    \label{tab:tseval-mt4ts}
    \begin{tabular}{l|cc}
    \toprule
    Model & \multicolumn{2}{c}{Top-1 Acc \%}\\
    & \multicolumn{1}{|c}{Top-100} & \multicolumn{1}{c}{Overall} \\
    \midrule
    TypeBERT & \multicolumn{1}{c}{48.92} & \multicolumn{1}{c}{28.07} \\
    GCBert-4TS & \multicolumn{1}{c}{87.22} & \multicolumn{1}{c}{63.42} \\
    CodeTIDAL5 & \multicolumn{1}{c}{\textbf{90.03}} & \multicolumn{1}{c}{\textbf{71.27}} \\
    \bottomrule
    \end{tabular}
\end{table}

\subsection{Efficacy of JoernTI}
\label{sec:pa}

The following section investigates the integration of CodeTIDAL5 into Joern as discussed in Section~\ref{sec:joernti}. 
We conduct partial program analysis, that is, no library dependencies are retrieved and integrated into the analysis and only application code is passed to Joern. 
This downstream task illustrates a less resource intensive program analysis task, where CodeTIDAL5 compensates for the missing external type dependencies. 
All experiments were conducted on an M1 MacBook Pro (2020), 16 GB RAM.

\subsubsection{Inference of Missing Types}
We infer types on 10 open-source JavaScript repositories from GitHub and manually review the inferred results.
The slices given CodeTIDAL5 to infer types for were ones where type information could not be easily resolved via the simple type propagation strategy already in place. 
We label the suggested types under the following four categories:
 \begin{enumerate}[noitemsep,topsep=0pt]
     \item Correct: The inferred type is an exact match with the actual type.
     \item Partial: The inferred type behaves similarly to the actual type or is its interface or supertype, e.g. \texttt{\_\_ecma.Request} versus inferring \texttt{NextApiRequest}.
     \item Useful: The actual type is not a defined type in the current context, but is plausible, e.g. \texttt{Object\{name: String, pass: String, email: String\}} versus inferring \texttt{User}.
     \item Incorrect: The inferred type is completely incorrect.
 \end{enumerate}
To justify the value of the \emph{partial} and \emph{useful} categories, we use examples from the manual review.
Consider two popular JavaScript REST frameworks: Express.js and Next.js. 
When defining routes, both accept handler functions defined with two parameters, where many tutorials and the official documentation often name these two parameters \texttt{req} and \texttt{res} for the request and response variables respectively. 
We find that CodeTIDAL5 may confuse the types from HTTP request frameworks such as Express.js \texttt{Request} with the Next.js \texttt{NextApiRequest}. 
Semantically speaking, both types perform the same roles, and we did not find CodeTIDAL5 confusing a response type with a request type. 
In these circumstances, we use the label \emph{partial}, since if the intention is to tag an HTTP response or request, then CodeTIDAL5 provides the correct subtype. 


For the \emph{useful} category, we describe a case in Project 1 where no classes are defined within the application and the developer simply uses the raw responses from the Postgres database responses. 
As CodeTIDAL5 is based on {CodeT5+}, which has some understanding of identifiers of commonly defined types, it suggests the non-existent type \texttt{Customer} for an identifier \texttt{customer}. 
Heuristically, we can consider a type of this name to hold sensitive Personally Identifiable Information (PII) such as a home address or email.

\begin{table}[tp]
    \caption{The results of manual reviewing 1093 type inference results from JoernTI. Selected projects can be found in the Appendix (cf.~\ref{tab:ossProjects}).}
    \centering
        \begin{tabular}{l|cccc}
        \toprule
        \multicolumn{1}{c|}{\multirow{2}{*}{Category}} & \multicolumn{4}{c}{Manual Labelling Results \%} \\
        \multicolumn{1}{c|}{} & Correct & Partial & Useful & Incorrect \\ 
        \midrule 
        Built-in & 76.60 & 0.00 & 1.25 & 22.14 \\
        User-Defined & 63.63 & 10.43 & 6.15 & 19.79 \\ 
        \midrule
        Overall & 72.10 & 3.57 & 2.93 & 21.41 \\ 
        \bottomrule
        \end{tabular}
    \label{tab:manualReviewTask}
\end{table}

\subsubsection{Improvement of Dataflow Analysis}
\label{sec:dfanalysis}
We conduct a case study in order to demonstrate how better type information may lead to enhanced analysis results in Joern.
Consider the following scenario: 
As security researchers, we are looking for potential database injections (CWE-943) in the context of AWS' DynamoDB~\cite{dynamodb}. This scenario is illustrated by our running example in Listing \ref{lst:codeAppendix}.
Knowing that the bodies of incoming events are a potentially attacker-controlled source, we develop the following Joern query:
\begin{lstlisting}[
    language=Scala,
    caption={Joern query to find a dataflow from an unsanitized request body from multiple frameworks to a DynamoDB query.}, % Captions are compulsory
    label={lst:query}
]
def src = cpg.identifier
  .typeFullName(".*(express.|NextApi|__ecma.)Request")
  .inFieldAccess.code(".*\\.body\\..*")
def sink = cpg.identifier
  .where(_.and(_.typeFullName(".*DocumentClient"), 
               _.argumentIndex(0)) // Receivers are at 0
  ).inCall.name("query")
sink.reachableBy(src)
\end{lstlisting}
The query looks for a dataflow from a special element, in this case parameters tainted by an object of types \texttt{\_\_ecma.Request}, \texttt{\_\_express.Request}, or \texttt{NextApiRequest}, into database query logic.
Without proper sanitization, this may lead to a NoSQL injection.
First, we create a CPG for the above-mentioned source code without the JoernTI backend.
Querying this CPG yields an empty list as the result, as no dataflow can be found without the missing type information.
Computing the CPG with JoernTI enabled, CodeTIDAL5 is able to correctly 
infer type \texttt{NextApiRequest} for the \texttt{req} variable.
Running the same query again on the new CPG, the dataflow is correctly identified, and we can continue analyzing the potential vulnerability.
As we compute the dataflow based on matches by type knowledge and interprocedural taint tracking, the query generalizes to diverse code no matter the syntactical structure.

\subsubsection{Results}
We show the results for type inference and manual inspection in Table \ref{tab:manualReviewTask}. 
Each entry is counted \emph{per-definition} and not \emph{per-occurrence} of the target object. 
We load a TypeScript declaration file for built-in types to filter inferences that violate the type constraints of the slice, and omit any \texttt{object}, \texttt{UNK} or \texttt{void} type inferences as unhelpful. 
End-to-end, the rate was 8.72 predictions per second.
We observe 72.10\% correctly inferred types, which matches our MT4TS benchmark.
It is noteworthy that certain types, such as JavaScript \texttt{Event} types, were often incorrectly classified due to little context within their usage.
For one code base, we were able to retrieve \texttt{Result} types from database queries which may contribute to more accurate sensitive source discovery. The model performed well on inferring user-defined types from libraries such as Express.js, Next.js, MongoDB, Postgres, and the AWS SDK, which is likely due to the popularity of these technologies. 
The investigated projects showed an average of 61.59\% typed nodes, including user annotations and inferences from the simple type propagation pass.
After inference with CodeTIDAL5, JoernTI on average contributes 8.58\% absolute more types on complex real-world software, inferring types at locations we otherwise would have no information for.

Concluding with our case study, we effectively demonstrate how and where SAST tools such as Joern benefit from additional type information.
The types inferred with CodeTIDAL5 directly lead to the discovery of a new dataflow, uncovering a potential NoSQL injection with high generalization potential.

\section{Related Work}
\label{sec:related}

\subsubsection{Neural Type Inference} 
All three models used in our evaluation (Section \ref{sec:eval}) make use of statistical approaches with deep learning models to infer types from code. 
In addition to these models for TypeScript, there are models designed to infer types for Python, such as HiTyper \cite{peng2022static} and TypeT5 \cite{typet5}. 
DeepTyper \cite{hell18} was a pioneering model in the direction of neural type inference for JavaScript, but was soon surpassed by the likes of LambdaNet \cite{lambdanet} and OptTyper \cite{pan20}. 
A concern in the direction of work towards statistical methods for type inference is the difficulty in teaching models to understand the logical constraints of code and type systems. 
While, in practice, it appears that these models work well in hybrid settings where a developer can validate the results \cite{voruganti2022flextype, typet5}, some approaches tried incorporating constraints directly.
Both LambdaNet and HiTyper incorporate type dependencies in the form of graphs and use type constraints as part of the learning and inference pipeline.
More recently, models such as TypeBert, DiverseTyper \cite{diversetyper}, TypeT5, and GCBert-4TS~\cite{mt4ts} used LLMs to achieve precise type inference.
As these models can take into account context and understand the language of naming strategies in code, with a large enough database to train from, these prove effective at predicting popular library types.
These approaches usually make no attempt to constrain the model itself, but instead prune poor predictions after the types are inferred. 
TypeT5, developed concurrently with our work, additionally uses static analysis to construct dynamic contexts for inference, somewhat similar in intuition to CodeTIDAL5.
It is notable, however, that these type dependency graphs in LambdaNet and HiTyper, as well as the usage graphs in TypeT5 do not appear to be language-specific and may generalize among other languages when trained on large enough corpora.


\subsubsection{End-to-End Statistical Type Inference for Developers}
FlexType \cite{voruganti2022flextype} is a Graph\-CodeBERT-based type inference plug-in with the aim to address the issue of being able to utilize this kind of model on a consumer-level laptop. 
The authors mention that models such as HiTyper and LambdaNet have large hardware requirements comparable to that of workstations or GPU servers. 
We share a common goal with CodeTIDAL5, i.e., to offer powerful type inference while requiring compute resources akin to a CI/CD runner or laptop. FlexType's downstream goal is to integrate into an IDE to enhance a developer's experience by inferring types during development, while ours is focused on recovering library types of interest which may be associated with sensitive data sources and sinks. 

\section{Conclusion}
\label{sec:concl}

We presented CodeTIDAL5, a neural type inference model based on CodeT5 that uses source code context as well as precise slices to query variable types in JavaScript/TypeScript. 
Additionally, we demonstrated the plug-and-play capability of CodeTIDAL5 using JoernTI, a queryable server for remote type inference, available as an open-source extension to Joern. 
We demonstrate the value of neural type inference on real-world partial-program analysis by recovering types for popular third-party libraries. 
Our experiments show that our approach leads state-of-the-art (SOTA) in the LambdaNet dataset for user-defined types by 6.31\% and is on-par with SOTA overall but exceeds SOTA by 7.85\% on the ManyTypes4TypeScript dataset.
The proposed system is able to boost Joern's existing type propagation, resulting in 8.6\% more typed nodes on real-world code bases and leading to improved downstream static analysis capabilities.

\subsubsection{Future Work}
While we demonstrated the capability of our neural type inference approach on JavaScript/TypeScript, the pipeline is language agnostic as it follows on the CPG directly. 
In future work, we expect it to be possible to reproduce this work on other Joern-supported languages with minimal modifications and to explore approaches for a single, unified model inferring types for multiple programming languages.

\subsubsection{Availability}
The source code of CodeTIDAL5 and JoernTI, as well as evaluation code and experiment data, is publicly available at: \\
\url{https://github.com/joernio/joernti-codetidal5}

\subsubsection{Acknowledgements}
The authors gratefully acknowledge funding from the European Union’s Horizon 2020 research and innovation programme under project
TESTABLE, grant agreement No. 101019206, from the German Federal Ministry of Education and Research (BMBF) under the grant BIFOLD (BIFOLD23B), the National Research Foundation (NRF), and Stellenbosch University Postgraduate Scholarship Programme (PSP).
We would also like to thank Kevin Jesse for help with the MT4TS dataset and models and the anonymous reviewers for the feedback on our work.


%
%
%
%

\newpage
\appendix
\section*{Appendix}
\label{sec:ossAppendix}

\begin{table}[htbp]
\centering
\caption{The GitHub subdirectories of each manually reviewed open-source web application or library. Notable technologies include React, Express, Chroma, MongoDB, Meteor, AWS, and Postgres.}
\begin{tabular}{l|l}
\toprule
\# & GitHub Subdirectory \\  
\midrule
1  & \href{https://github.com/skirupa/Bank-Management-System}{/skirupa/Bank-Management-System}                                  \\
2  & \href{https://github.com/najathi/shopping-app-mongodb}{/najathi/shopping-app-mongodb}                                      \\
3  & \href{https://github.com/Hamidreza-khushab/express-server}{/Hamidreza-khushab/express-server}                              \\
4  & \href{https://github.com/meesont/node-house-scoring-system}{/meesont/node-house-scoring-system}                            \\
5  & \href{https://github.com/Dynatrace/AWSDevOpsTutorial}{/Dynatrace/AWSDevOpsTutorial}                                        \\
6  & \href{https://github.com/Hthe-scan-project/vulnerable-app-nodejs-express}{/Hthe-scan-project/vulnerable-app-nodejs-express}\\
7  & \href{https://github.com/qrohlf/trianglify}{/qrohlf/trianglify}                                                            \\
8  & \href{https://github.com/themeteorchef/base}{/themeteorchef/base}                                                          \\
9  & \href{https://github.com/jaredhanson/passport}{/jaredhanson/passport}                                                      \\
10 & \href{https://github.com/OWASP/NodeGoat}{/OWASP/NodeGoat}                                                                  \\
\bottomrule
\end{tabular}

\label{tab:ossProjects}
\end{table}

\bibliographystyle{splncs04}
\bibliography{references}

\end{document}